\documentclass[letterpaper]{article}
\usepackage{flairs}
\usepackage{times}
\usepackage{helvet}
\usepackage{courier}
\usepackage{graphicx}
\begin{document}
%
\title{Visual Question Answering Using Semantic Information from Image Descriptions}

\author{Tasmia Tasrin\thanks{These authors have contributed equally.} \\
University of Kentucky\\
{\tt\small tta245@uky.edu}
\And 
Md Sultan Al Nahian\footnotemark[1]\\
University of Kentucky\\
{\tt\small mna245@uky.edu}
\And
Brent Harrison\\
University of Kentucky\\
{\tt\small harrison@cs.uky.edu}

}
\maketitle
\begin{abstract}
\begin{quote}
 In this work, we propose a deep neural architecture that uses an attention mechanism which utilizes region based image features, the natural language question asked, and semantic knowledge extracted from the regions of an image to produce open-ended answers for questions asked in a visual question answering (VQA) task. 
 The combination of both region based features and region based textual information about the image bolsters a model to more accurately respond to questions and potentially do so with less required training data.
We evaluate our proposed architecture on a VQA task against a strong baseline and show that our method achieves excellent results on this task. 
\end{quote}
\end{abstract}

\section{Introduction}

Visual Question Answering (VQA) is a task in which a system provides natural language answers to questions concerning an image. 
This is typically accomplished using deep learning systems that extract textual features from the question and image features from the related image. 
One difficulty that arises with this approach is that there is very little text signal that networks can use to derive semantic information in the image. 
This means that either 1) Large amounts of question and answer data must be gathered for ML techniques to learn effectively or 2) Approaches will struggle to draw a connection between image features and text semantics.

To overcome these shortcomings, we propose to augment a deep learning architecture that utilizes neural attention with additional, external knowledge about the image. 
This type of approach has been used in the past~\cite{imgcap,askme,genpara,qcap}; however, our work seeks to take advantage of a different form of knowledge. 
Our resulting network, which we call the Visual Question Answering-Contextual Information network (VQA-CoIn), improves upon past work by extending it to incorporate semantic information extracted from every regions an image via image descriptions.
We hypothesize that this will help bridge the gap between image features and natural language text, which should improve network understanding.

In our proposed model, we have extend the work in \cite{NIPS2018_7429} by incorporating semantic information of an image's regions in addition to the image features and questions.
To evaluate the effectiveness that this additional information has, We have used the VQA v2.0 \cite{VQA2} dataset to train and evaluate our architecture. 
For generating semantic information for each image of the dataset, we utilize Densecap\cite{Densecap} to extract all possible image regions and produce image captions for them. 
We preprocess these captions to extract the important words out of them as we want to make sure that the information limits to a fixed length and the deep learning network used to encode these words focus on the meaningful texts. 
For evaluation of our network, we use accuracy as the evaluation metric like all prior works. 
To measure the test accuracy score, we use VQA challenge hosting site EvalAI where we submit the generated results and the site measures the scores for 3 categorical questions and overall test accuracy for the test split data. 
In addition, we also evaluate how well our techniques scales with data compared to other techniques by testing with different percentages of the training dataset. 



\section{Related Works}

The introduction of the VQA v1.0 and VQA v2.0 datasets \cite{VQA,VQA2} have drastically accelerated research in this area. 
As a result, some interesting works like \cite{stacked,dasetal2016,showask,Anderson2017up-down,tips} have been proposed for the advancement in VQA. 
For instance, while the work in \cite{stacked} focuses on putting stacked attention on images, \cite{Anderson2017up-down} has assigned bottom-up attention to figure out the regions and then used top-down mechanism to determine the important features. 
Both \cite{stacked} and \cite{Anderson2017up-down} are the winners of the VQA challenge 2016 and 2017 respectively. 
The region based features introduced by \cite{Anderson2017up-down} have been utilized by several approaches \cite{NIPS2018_7429,py} including ours. 
In the 2018 VQA challenge, authors improved upon the 2017 winner by changing learning schedule, fine tuning the model, and using both grid level features and region features of images \cite{py}. 
But recently a research work has revisited grid based features of VQA and used them in end-to-end training \cite{grid}, which has produced strong results.

There has been prior work that merges elements of image and text for the VQA task.
For example, Kim and Bansal utilized both paragraph captions and object properties described using sentences as prior knowledge to aid in the VQA task on the visual genome dataset \cite{genpara}.
Wu \textit{et al.} proposed a free-form VQA model where internal textual representations of an image are merged with textual information sourced from a knowledge base~\cite{askme}. 
They claim that by using this method, a VQA system can answer both complex and broad questions conditioned on images.
In work by Wu, Hu, and Mooney \cite{qcap}, captions are generated from questions and are used as an input to the neural network. 
The model we propose in this work focuses on using region-based contextual information as we believe that performance of a VQA model can be bolstered if more knowledge about an image is provided through both region-based visual features and textual information.
These semantic texts about images can better help our model to bridge the gap between images and natural language questions as well.

\section{Method}
Given an image and associated question, the aim of our proposed model is to produce an answer of the question utilizing the salient image features and semantic information of the image. 
In all upcoming sections, semantic information will be approached as SI.
We name our model as VQA-contextual information (VQA-CoIn) model and following this, in the rest of the paper, we will address the method by VQA-CoIn model. 
Through our network, we emphasize that adding contextual information of the image can help a VQA agent to learn more about the content of the target image to answer relative questions about the image. 

In our proposed architecture, there are three different modules which make an end to end deep learning architecture for the Visual Question Answering task. 
The first module:\textit{Input Encoder}, where visual features are extracted using Faster r-CNN\cite{NIPS2015_5638} method and textual features like natural language questions and semantic knowledge of the images are embedded using gated recurrent units (GRUs) \cite{GRU}. 

The next process in our model involves attending to images and SI of images works using a bilinear attention mechanism\cite{NIPS2018_7429} conditioned on question embedding vectors. 
We also apply self attention on questions to learn the importance of the words residing in the given questions.   
The third component is the \textit{classifier} which predicts candidate answers using the concatenated vector produced from the sum pooling of the three output vectors of the previous two modules. 
Figure 1 shows an overview of our VQA-CoIn model.
We will discuss each module of our architecture in greater detail below. 

\begin{figure}[ht]
\centering
\includegraphics[width=\columnwidth]{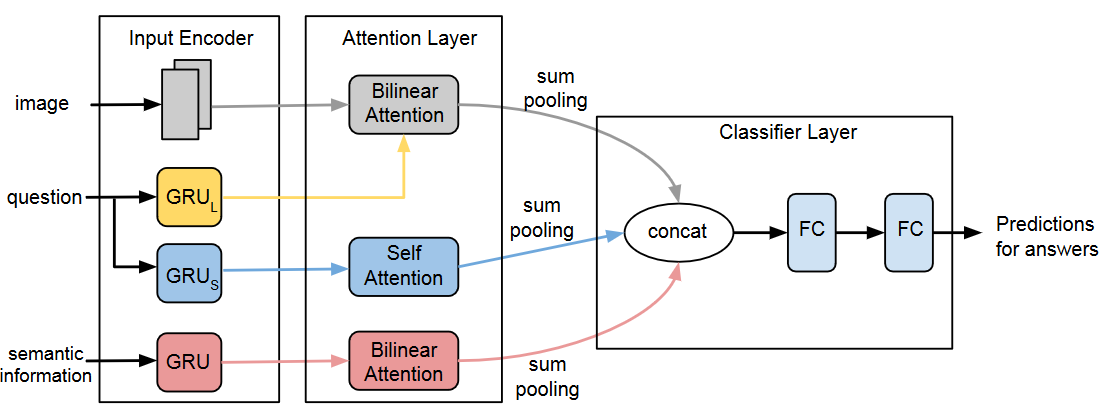}
\caption{Overview of the proposed VQA-CoIn architecture. 
}
\label{fig_architecture}
\end{figure}

\subsection{Input Encoder}

The Input Encoder takes an image, associated question, and the SI of the image as inputs and produces three embedding vectors, one from each of the inputs.
For the image features, we use the pre-trained bottom-up attention features which were generated in \cite{Anderson2017up-down}. 
They used Faster r-CNN algorithm \cite{NIPS2015_5638} with ResNet-101 to train the model.
Adaptive number of features $f_i$ per image is considered to generate the vector representation $f_i \times d_i$ for each image where $d_i$ is the image feature size.

Our model uses semantic information (SI) extracted from the image as external knowledge to learn more about the image. 
To encode these semantic features, each word is embedded using pre-trained tf-idf word embeddings. 
Then the embedded word vectors are propagated through GRU cell which gives us hidden vector for each corresponding word. 
The hidden vectors are used in the attention layer to create the encoding vector of the semantic features.

\begin{figure*}[ht]
\centering
\includegraphics[scale=1.7]{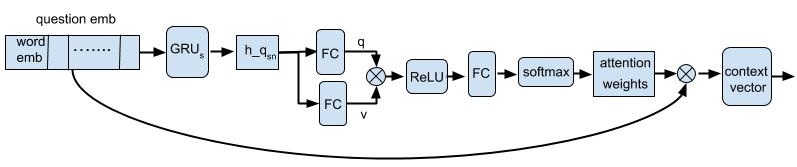} 
\caption{A detailed architecture of self attention mechanism of questions with smaller hidden states $h\_q_{sn}$. Here, n defines the number of words in a question. These hidden states are passed into two FC layers and then multiplied to create a joint representation. This joint representation is  forwarded through ReLU and FC layers to get the attention weights. The question embedding vector is element-wise multiplied with the attention weights to achieve the context vector. }
\label{fig_att}
\end{figure*}

The final input for our task is the question related to the image which is embedded twice with two different hidden vector sizes in our model.
We do this because we realize from our experiments that questions with large hidden vectors can contain more information from image while attending them.
But in the case of prioritizing information from its own features, hidden layers with smaller dimensions can perform this task more effectively.
In Figure \ref{fig_architecture}, yellow is used for questions with large vector sizes and blue is for questions with smaller dimensions. 
As with the SI embedding, we use the same pre-trained tf-idf word embeddings to embed each word of a question. 
The obtained word embedding vector is then passed through two separate GRU cells \cite{chung2014empirical} represented as GRU\textsubscript{l} and GRU\textsubscript{s} for larger and smaller hidden state respectively in Figure \ref{fig_architecture}. 
The GRU cells have $n_q$ numbers of hidden states $h\_q_l$ and $h\_q_s$.
Here, $n_q$ is the word count in the question and $h\_q_l$ and $h\_q_s$ denote the hidden states from GRU\textsubscript{L} and GRU\textsubscript{S} respectively.
Hidden states of both GRU cells are passed to the attention layer to generate the context vector of the questions.

\subsection{Attention Layer}
In the attention layer, we employ an attention mechanism to find out the importance of different parts of the input sequence based on a query vector. We use two different attention mechanisms on the input vectors: 1. Self-Attention, applied to question embedding vector and 2. Bi-Linear Attention, applied to image embedding, question embedding and SI embedding vectors.

\subsubsection{Self-Attention}
Self-attention is an attention mechanism where relations among different parts of a sequence are computed using the same sequence as query. 
In our proposed architecture, we apply self-attention on a question to figure out the internal relations among the words of a question.
For example, in a question like `what is the color of the bus?', invoking self-attention on itself would enable the model to identify that the words `color' and `bus' are interrelated and should be more emphasized to learn about the question.

Figure \ref{fig_att} shows the detailed architecture of the self-attention module.
We implement the self-attention mechanism inspired by the idea of multi-headed attention which are featured in many transformer architectures \cite{NIPS2017trans}.
In this process, we take into account all of the hidden states of the GRU\textsubscript{S} instead of the final hidden state as RNNs have a tendency to forget the information encountered in the early steps of the sequence. 
All of the hidden states are passed through two fully connected (FC) layers which generate two vectors, a query vector $q$ and a value vector $v$. 
In each FC layer, weight normalization and ReLU activation are performed on the input vectors.
Then the resultant vectors $q$ and $v$ are multiplied together to create a new context vector. 
Here, the multiplication operation is the element-wise multiplication (Hadamard product) of the vectors.
The new context vector is forwarded to another FC layer followed by a softmax layer to generate the attention weights of the input question embedding. 
Afterwards, these weights are multiplied with the initial question embedding to construct the final context vector $c\_q_s$.
This final question context vector represents the prioritized words in the input question sequence. 
As a next step, $c\_q_s$ is used to put bilinear attention \cite{NIPS2018_7429} on the semantic concepts of the images.


\subsubsection{Bilinear Attention}
Bilinear attention is usually applied on two inputs with multiple channels so that the two input channels decrease their dimensionality concurrently.
We adopt this attention mechanism from Bilinear attention networks \cite{NIPS2018_7429}. 
In our case, we have two input groups to apply the attention: one group is the combined group of the image and question and the other is the combined group of the SI and question context vector.
In the attention procedure, at first an attention map is generated using image features conditioned on given questions embedded using GRU\textsubscript{L}.  
Similar to \cite{NIPS2018_7429}, this attention map is then run through eight glimpses. 
In each glimpse, a vector representation from the image and question is produced using the bilinear attention map. 
Next, with this representation, for every glimpse, we keep integrating the resultant vectors of the residual learning network and counter module \cite{count}.  
As a result, at the last glimpse, we get a final output vector $b\_c_v$.

For the input group of SI and the question context vector $c\_q_s$, we similarly produce an attention map using the two input vectors.
But unlike the image-question input group, we use one glimpse on the attention map of SI-question group to generate a vector.
Also, we element-wise add the context vector $c\_q_s$ of question with the resultant vector from the glimpse. 
This creates an output vector $b\_c_{si}$.

\subsection{Classifier Layer}

$b\_c_v$, $b\_c_{si}$ and $c\_q_s$ are the inputs of our classifier layer. 
The sum pooling of these three input vectors are concatenated as the next step of the classifier.
And the concatenated vector is then redirected to two FC layers to gather predicted answers for the questions.
In FC layers, we use ReLU as the activation function and the output dimension is set to the number of unique answers.
We have selected these answers that appear at least 9 times for the distinct questions in the training dataset.

\section{Experimental Setup}

In this section, we are going to have a detailed discussion about the implementation procedure and experimental setup for VQA-CoIn model. First, we discuss the dataset we use for our task and then the preprocessing of our additional prior semantic knowledge. 
We will then outline the network parameters for our proposed model that we use in the experiments. 
To evaluate our method VQA-CoIn, we have used the available VQA challenge guidelines. 
We use BAN-8 \cite{NIPS2018_7429} as our baseline. 
And we compare our validation and test scores with bottom-up attention model \cite{Anderson2017up-down} as well.

\subsection{Dataset}

We evaluate our proposed model on VQA v2.0 dataset. We use the provided train/validation split of the dataset to train our network. 
In the training dataset, more than 400k questions and 82k images and in the validation split, 200k questions and 40k images are available.
Though we are utilizing full dataset, our model is trained to learn from the selective answers from the train split.
Recall that these are chosen, because they appear as answers at least 9 times for the unique questions of the split. 
The number of these selective answers is 3,129.
The full test split of the dataset has around 82k images and 440k relevant questions on which we test our network.
As VQA task is an open challenge, the ground truth answers for the questions in the test dataset are not available and cannot be compared with.

\subsection{Preprocessing}

As we have mentioned, our method exploits the semantic concepts of images available in the dataset as an input of our architecture.
To generate this information, we use Densecap \cite{Densecap}, an image captioning model.
For each image, Densecap generates a variable number of captions.
We have found that after a certain number of generated captions, the generated information tend to be duplicates. 
For example, Densecap has generated both `man wearing a hat' and `a man wearing a hat' captions for an image. 
To avoid these duplicate words, we have removed a sentence which has at least 80\% similarity with any previous selected sentence.

After discarding the similar sentences,  from the resultant list, first 10 sentences are taken and preprocessed. 
As preprocessing steps, we tag the words of each sentence with the NLTK part-of-speech tagger and then get rid of the stop words such as `the', as well as any preposition and auxiliary verbs from the sentences. 
Afterwards, the remaining words are gathered in a list which we have used as the SI for the respective image.

\begin{table}
\begin{center}
\begin{tabular}{|l|c|c|}
\hline
Scale\% & VQA-CoIn & BAN \\
\hline\hline
25 & \textbf{54.84} & 54.09 \\
50 & 61.76 & \textbf{62.42} \\
75 & \textbf{65.08} & 64.92\\
\hline
\end{tabular}
\end{center}
\caption{Validation accuracy after training VQA-CoIn with different scales of train split. }
\end{table}

\subsection{Network Parameters}

We consider our image feature size $d_i$ as 2048 and the number of features $f_i$ as variant between 10 to 100 per image. 
For word embedding, pre-trained GLoVe vectors of size 300 have been used. 
As we mention in section 3.1, questions are embedded twice in our architecture.
Question embeddings which are used for attending image features utilize GRU cells with a dimension of 1024 and the question features utilized for self-attention and external knowledge prioritization consists of a 512 sized vector.
The maximum word length $n_q$ for any embedded question in the proposed model is 14. 
To embed the semantic concept of an image, we fix the size of the GRU units to 512.
The additional semantic knowledge about an image can consist of maximum 40 words. 
For training, we find that 18 epochs are enough to sufficiently train the network. 
We have used the batch size of 180 for training and testing the dataset.
The Adamax optimizer is used to optimize the classifier and dropout value of the classifier is set to 0.5 while FC layers have dropout of 0.2.

\subsection{Baselines}

BAN-8 \cite{NIPS2018_7429} is our baseline architecture. 
To compare with the baseline model, we have re-trained the BAN model\cite{NIPS2018_7429} from their github repository to reproduce the results.
A note to mention, we deploy some changes to BAN model while reproducing it.
First, we use batch size 180 to run their model as we deploy for our VQA-CoIn model. 
Second, we have omitted the effect of data augmentation of visual genome dataset from BAN model as we have not employ any data augmentation.
The reason of making these changes is to appropriately compare our model with the baseline model.

\subsection{Evaluation Criterion}

For VQA tasks, question accuracy is the preferred evaluation metric. 
In this section, we are going to illustrate the evaluation process we have followed.
Like previous VQA approaches, we have computed accuracy to decide how our architecture is performing to figure out the correct answers for a given question using both image features and contextual information about the image.
According to the employed dataset, validation and test accuracy scores are calculated. 
Validation accuracy is measured by comparing against the ground truth answers available in the validation data split.
And to obtain the test score, we have generated answers for the questions in the test split based on our model and submitted the results in the VQA challenge hosting site EvalAI. 
It provides scores for each category of questions which are already defined in the dataset.

\begin{table}
\begin{center}
\begin{tabular}{|l|c|}
\hline
Method & Validation Score \\
\hline\hline
bottom-up & 63.20 \\
BAN-8 & 66.28 \\
VQA-CoIn(I+Q) & 66.03 \\
VQA-CoIn(I+Q+SI) & \textbf{66.33}\\

\hline
\end{tabular}
\end{center}
\caption{Validation scores computed on the full VQA v2.0 dataset for bottom-up model, BAN-8 and our architecture.  }
\end{table}

\begin{center}
\begin{table}
\begin{tabular}{ |p{1.6cm}|p{.8cm}|p{0.9cm}|p{0.8cm}|p{0.8cm}|p{1cm}|}
\hline
Method & yes/no & number & other & overall & test-dev \\
\hline\hline
bottom-up & - & - & - & 65.67 & 65.32 \\
BAN-8 & \textbf{83.61} & 50.45 & 58.12 & 67.75 & 68.07   \\
VQA-CoIn & 83.57 & \textbf{50.91} & \textbf{58.33} & \textbf{67.88} & \textbf{68.2} \\
\hline
\end{tabular}
\caption{Comparison of Test-standard scores among VQA-CoIn, BAN-8 and bottom-up model. VQA-CoIn are trained using VQA v2.0 train and validation splits and tested on test split. Note that all 3 models are single models. 
}
\end{table}
\end{center}

\begin{table*}
\begin{center}
\begin{tabular}{ |p{2.8cm}|p{3.8cm}|p{3.8cm}|p{3.8cm}|}
 \hline
 &
  \includegraphics[scale=0.22]{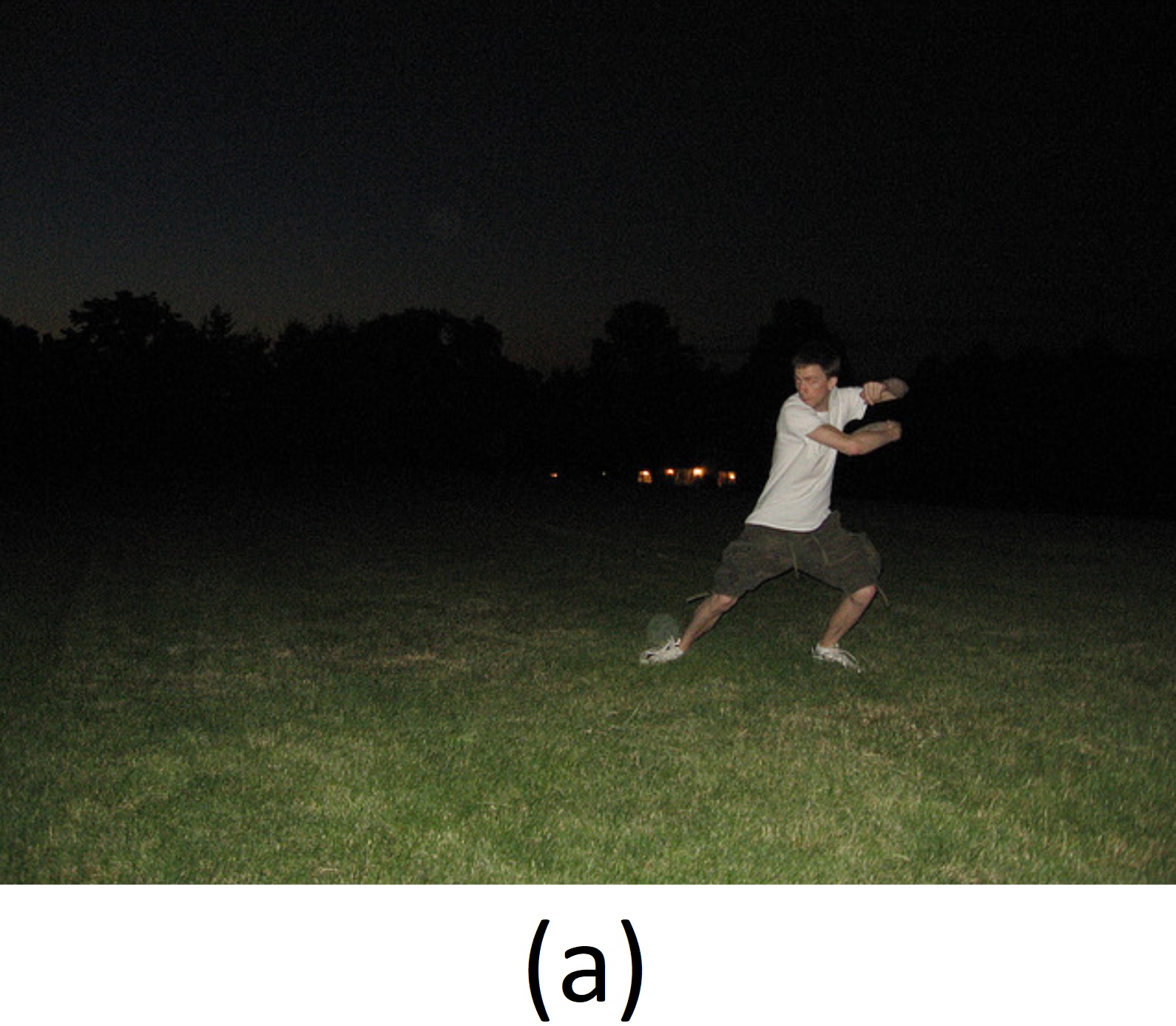} 
  &  \includegraphics[scale=0.3]{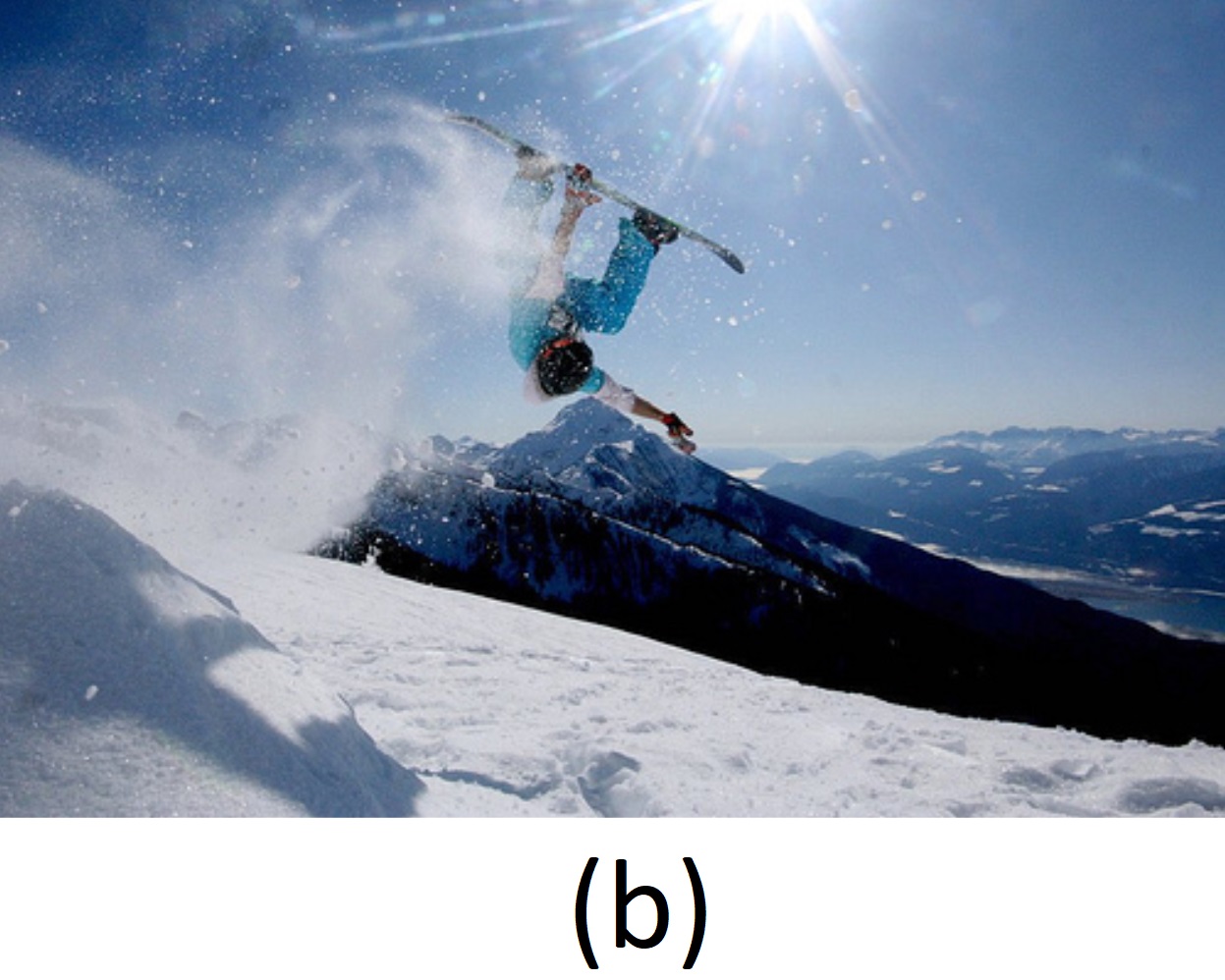} & \includegraphics[scale=0.23]{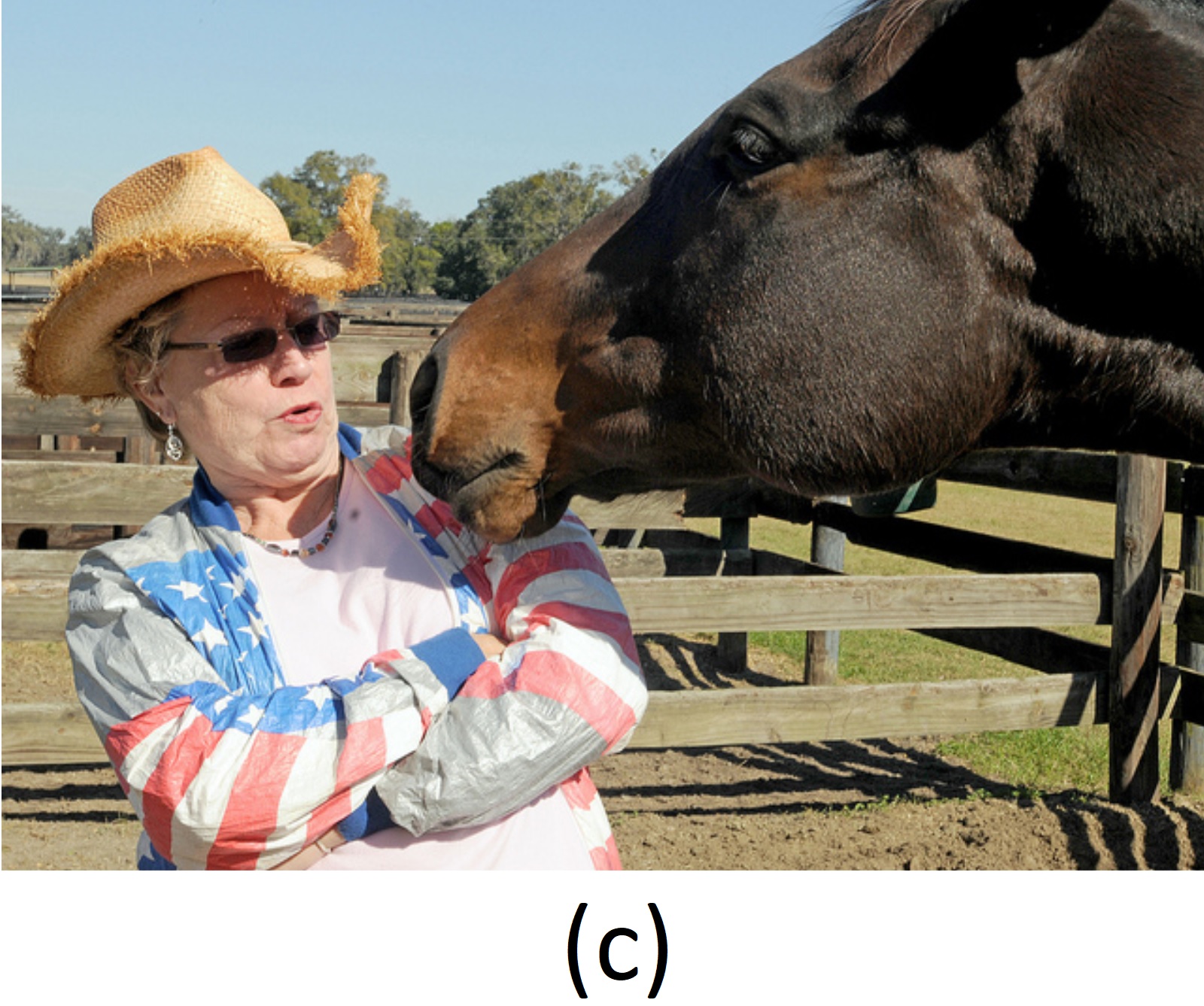}\\
  
 \hline
 question & What is he doing at night? & What sport is the man participating in?  & What color is the man wearing? \\
 \hline
 semantic info & 
 `man \textbf{playing frisbee}', `green grass field', `man wearing white shirt', `sky clear', `man wearing shorts', `man short hair'
 & `large blue sky', `person skiing', `man wearing black jacket', `person \textbf{snowboarding}', `snow covered mountain'
 & `black and white cow', `man wearing hat', `hat man', `man and woman sitting bench', 
 `\textbf{red and white striped shirt}'
 \\
 \hline
 ground truth answer & playing frisbee & snowborading & red, white and blue \\
 \hline
 VQA-CoIn answer & playing frisbee & snowboarding & red and white\\
 \hline
 BAN-8 answer & playing  & skiing & white\\
 \hline
\end{tabular}
\caption{Comparison of answers generated for questions and images from the validation data by VQA-CoIn and BAN-8 models. }\label{table:story}
\end{center}
\end{table*}

\section{Results \& Discussion}

We have considered BAN-8 \cite{NIPS2018_7429} model as our baseline and compared our results with it's single model validation and test scores. 
Unlike other approaches, we also execute an experiment in which we check the validation accuracy while a model is trained with different scales of training data.
Through this investigation, we want to figure out whether our model can learn and generate precise answers for the validation questions though it has been trained on different sets of train dataset.  

\subsection{Quantitative Results}
Table 1 demonstrates the results for our data scaling experiment.
We perform this experiment using on our VQA-CoIn model and our baseline model. 
We find that for one fourth and three fourth of the dataset, when we train our model, it is capable of functioning better than BAN-8 model. 
But while trained on 50\% of the training split, BAN model perform better than VQA-CoIn. 
The reason behind this could be, as we are enforcing contextual information of the images generated by a pre-trained model in our method, some of these information may not carry knowledge related to the question to answer it correctly.
This observation can lead our study to further investigation by producing and invoking SI using other pre-trained models in future.
We still feel that this gives strong evidence that our approach can better utilize small amounts of data when compared to state-of-the-art approaches. 
Through Table 2, we estimate the validation score of our model for the whole dataset with the state-of-the-art VQA models.  
The validation score on the \textit{VQA-CoIn (I+Q)} row of the table portrays the importance of the region based SI of images we employ through our model. 
While SI along with image(I) and question(Q) as inputs are given to the network, VQA-CoIn outperforms the state-of-art baseline architectures in terms of accuracy. 

In order to receive scores for the test set, we submitted the results  produced by our model to the VQA competition using EValAI. 
We also submit the reproduced answers of BAN in the same site to find out and compare the test-dev and test-standard scores with ours.
According to the results returned, displayed in Table 3, We can observe that VQA-CoIn has outperformed BAN and  bottom-up\cite{Anderson2017up-down} in test-dev and test-standard challenges.
If we consider each category of questions for BAN-8 and VQA-CoIn models, we can see that VQA-CoIn network has surpassed the scores of BAN-8 for `number' and `other' categorical questions. 
For `yes/no' category of questions, BAN has performed better than ours. 
We feel that these results are significant, especially our performance on the `other' category. 
To answer any question from `other' category, a model needs to understand more complex relation among the contents of an image (where to search for an answer). 
SI provides support behind this logic and helps our model to generate more accurate answers than our baseline models. 


\subsection{Qualitative Results}

After the quantitative comparison of our and two state-of-the-art models, we do a qualitative contrast between VQA-CoIn and BAN-8 using the data of validation split.
This is not meant to be a formal evaluation, but mainly meant to provide additional context to the results that our approach gives compared to our baselines. 
Table \ref{table:story} represents the contrast. 
We have image, question and SI for each of three examples. 
The human annotated ground truth answers for the examples are also added so that the answers generated by both of the models can be compared with it. 
From the table, we can see that for image (a) and (b), our model generates correct answers.
For the same images, BAN model generates answers that are very close to the answers from the dataset, but not accurate. 
Here, the reason of the success of our model is both image features and SI for images.
The answers for the questions are already available in the SI. 
For ease of reading, we bold the texts on the row named as semantic info in the table.
Now, if we match answers for image (c) of both models, answers are not exact to the ground truth answers.
Our model is able to detect only two colors using both image features and SI (bold texts in semantic info row under image (c)) available for the input question. 
So, VQA-CoIn chooses these two colors as answer.
It also means that, if better SI is used, our model can generate more correct answers. 


\section{Conclusion}

In this paper, we have proposed a novel VQA architecture, VQA-CoIn, which incorporates contextual information of every possible region of an image to understand and represent features of an image with already available textual information about it. 
Our motivation behind incorporating SI in the form of natural language descriptions is to better enable ML models to bridge the gap between the questions being asked and the image itself. 
We also hypothesize that this should result in better data scaling, and enable ML models to perform well with less data. 
We have compared our VQA-CoIn model with two state-of-the-art models and showed that our model performs better than those models both in terms of raw accuracy, and in terms of scaling performance. 
As our future work, we intend to apply our model to build applications to help visually impaired to guide them with human understandable texts. 
We also have a plan to do human evaluation of the results we achieve.

\bibliography{flairs_vqa}

\end{document}